\documentclass[conference]{IEEEtran}
\IEEEoverridecommandlockouts
\usepackage{cite}
\usepackage{amsmath,amssymb,amsfonts}
\usepackage{algorithmic}
\usepackage{graphicx}
\usepackage{textcomp}
\usepackage{xcolor}
\def\BibTeX{{\rm B\kern-.05em{\sc i\kern-.025em b}\kern-.08em
    T\kern-.1667em\lower.7ex\hbox{E}\kern-.125emX}}

\usepackage{url}
\usepackage{multicol}
\usepackage{multirow}
\usepackage{color}
\usepackage{hhline}
\usepackage{pifont}
\usepackage{subcaption}
\usepackage{multirow}
\usepackage{tikz}
\usetikzlibrary{decorations.pathreplacing,calc}
\newcommand{\tikzmark}[1]{\tikz[overlay,remember picture] \node (#1) {};}

\newcommand\Alpha{\mathcal{A}}
\DeclareMathAlphabet{\mathpzc}{OT1}{pzc}{m}{it}

\usepackage[ruled,vlined]{algorithm2e}
\SetKw{KwBy}{by}

\definecolor{mynicegreen}{RGB}{11,102,35}

\newcommand{\model}{{\textsc{MRCo}}}

 \newcommand{\squishlist}{
	\begin{list}{$\bullet$}
		{ \setlength{\itemsep}{0pt}
			\setlength{\parsep}{3pt}
			\setlength{\topsep}{3pt}
			\setlength{\partopsep}{0pt}
			\setlength{\leftmargin}{1.5em}
			\setlength{\labelwidth}{1em}
			\setlength{\labelsep}{0.5em} } }
	
	\newcommand{\squishlisttwo}{
		\begin{list}{$\bullet$}
			{ \setlength{\itemsep}{0pt}
				\setlength{\parsep}{0pt}
				\setlength{\topsep}{0pt}
				\setlength{\partopsep}{0pt}
				\setlength{\leftmargin}{2em}
				\setlength{\labelwidth}{1.5em}
				\setlength{\labelsep}{0.5em} } }
		
		\newcommand{\squishend}{
	\end{list}  }
\begin{document}

\title{Reducing and Exploiting Data Augmentation Noise through Meta Reweighting Contrastive Learning \\ for Text Classification}

\author{\IEEEauthorblockN{Guanyi Mou\textsuperscript{\textsection}}
\IEEEauthorblockA{Worcester Polytechnic Institute \\
100 Institute Rd,
Worcester, MA, 01605 \\
gmou@wpi.edu}
\and
\IEEEauthorblockN{Yichuan Li\textsuperscript{\textsection}}
\IEEEauthorblockA{Worcester Polytechnic Institute \\
100 Institute Rd,
Worcester, MA, 01605 \\
yli29@wpi.edu}
\and
\IEEEauthorblockN{Kyumin Lee}
\IEEEauthorblockA{Worcester Polytechnic Institute \\
100 Institute Rd,
Worcester, MA, 01605 \\
kmlee@wpi.edu}
}
\IEEEoverridecommandlockouts
\IEEEpubid{\makebox[\columnwidth]{978-1-6654-3902-2/21/\$31.00~\copyright2021 IEEE \hfill} \hspace{\columnsep}\makebox[\columnwidth]{ }}
\maketitle
\IEEEpubidadjcol
\begingroup\renewcommand\thefootnote{\textsection}
\footnotetext{Equal contribution.}
\endgroup

\begin{abstract}
Data augmentation has shown its effectiveness in resolving the data-hungry problem and improving model's generalization ability.  
  However, the quality of augmented data can be varied, especially compared with the raw/original data.
  To boost deep learning models' performance given augmented data/samples in text classification tasks, we propose a novel framework, which leverages both meta learning and contrastive learning techniques as parts of our design for reweighting the augmented samples and refining their feature representations based on their quality. As part of the framework, we propose novel weight-dependent enqueue and dequeue algorithms to utilize augmented samples' weight/quality information effectively. Through experiments, we show that our framework can reasonably cooperate with existing deep learning models (e.g., RoBERTa-base and Text-CNN) and augmentation techniques (e.g., Wordnet and Easydata) for specific supervised learning tasks. Experiment results show that our framework achieves an average of 1.6\%, up to 4.3\% absolute improvement on Text-CNN encoders and an average of 1.4\%, up to 4.4\% absolute improvement on RoBERTa-base encoders on seven GLUE benchmark datasets compared with the best baseline.
  We present an in-depth analysis of our framework design, revealing the non-trivial contributions of our network components. Our code is publicly available for better reproducibility.~\footnote{\url{https://github.com/bigheiniu/BigData_MRCo}}
\end{abstract}

\begin{IEEEkeywords}
text classification, data augmentation, meta learning, contrastive learning
\end{IEEEkeywords}

\section{Introduction}
\label{sec: intro}

Data augmentation has demonstrated its effectiveness among lots of works, from image classification~\cite{perez2017effectiveness} to text classification~\cite{bayer2021survey}. It not only reduces the human efforts in labeling samples but also improves models' generalization ability~\cite{perez2017effectiveness}. It works by ``synthesizing new data from existing training data''~\cite{anaby2020not}. Specifically, it usually applies label-preserving transformations to the raw/original training data to create more samples and enrich the training dataset.

However, the quality and the contribution of augmented instances/samples\footnote{We use terms of ``instance'' and ``sample'' interchangeably.} are not always guaranteed~\cite{yi2021reweighting}. Augmented samples can vary at a macro level in terms of their quality because augmentation methods have different ways to produce augmented samples. Some augmentation methods can be more effective than the others in either a certain task or generally various tasks~\cite{shu2019meta}. Augmented samples can also vary instance-by-instance at a micro level -- even the same augmentation method may not be effective on all raw instances. What is more, augmented samples generated by the same augmentation method and the same raw/original instance can also contribute differently because of their data quality difference.
For example, in the binary sentiment classification~\cite{socher2013recursive}, randomly deleting neutral words may not cause a significant impact on a review's sentiment, but deleting a negation word is likely to flip its label. 

In search of finding proper ways to reduce the noise\footnote{We treat noise as the negative impact of low-quality augmented samples.} from augmented instances, researchers also proposed solutions at macro and micro levels. At a macro level, researchers try to identify the effective augmentation methods and then apply them to the raw data. The identification of effective augmentation methods can be non-differentiable or differentiable.
The non-differentiable approach treats different augmentation methods as a hyperparameter and selects the best one or groups them based on validation performance~\cite{qu2020coda}. However, it may require domain knowledge to design or construct the candidate augmentation methods' pool~\cite{tang2020online}. In addition, these augmentation methods may also include additional hyperparameters to tune.
It will impede the process of generating high-quality augmented samples.
For example, CoDa~\cite{qu2020coda} tried various combinations of stacking augmentations, which is in practice, trying out various sequential binary valued hyperparameters.
On the other hand, the differentiable approach considers optimizing the augmentation method and downstream task model together through back-propagation on defined objective functions~\cite{tang2020online,hu2019learning,cubuk2019autoaugment}. Although it does not require human prior knowledge to design augmentation methods,  it constrains the potential augmentation methods to be differentiable. Thus, traditional heuristic augmentations with discrete transformation operations cannot be fitted in such a setting.
%
Moreover, such an approach would require additional backward propagation to update the parameters of augmentation methods. The whole process can be time-consuming and non-trivial to identify the suitable augmentation methods for different tasks.
Thus the scope of augmentations is limited.

At the micro-level, researchers attempt to identify effective augmented samples. They consider the augmentation methods as black-box. Given the augmented samples, most of the works utilize heuristic rules to filter instances or training a reweighting module to reweight augmented instances' loss. In heuristic metric filtering methods, people introduce prior knowledge/domain expertise related to the tasks (e.g., readability). However, human expert knowledge can be suboptimal in many cases. In sample reweighting approaches~\cite{yi2021reweighting,shu2019meta,ren2019learning,zhou2020metaaugment}, they assign an importance score for each augmented sample and minimize the reweighted loss on these samples. One of the challenges in the reweighting approaches is how to optimize the reweighting module and downstream task model together, as directly minimizing vanilla reweighted loss makes the reweighted module collapse by setting the weights close to 0 to minimize the loss. Many reweighting approaches~\cite{shu2019meta,ren2019learning,zhou2020metaaugment} consider utilizing the meta learning to convert the optimization into the bilevel optimization. Other works like Yi et al.~\cite{yi2021reweighting} add regularization to encourage the weight towards uniform distribution to avoid the collapse.
Ideally, the reweighting model will assign small weights to low-quality samples, thus neglecting the contribution of these samples.

In a nutshell, macro-level approaches of reducing the augmentation noise cannot be applied to all the augmented samples.
In micro-level approaches, although they are accessible for all the augmentation methods, they will ignore the potential contribution of filtered out samples.
We believe a better approach shall incorporate as many augmentation methods and as many augmented samples as possible (no seemingly obvious upper limit in capacity), with minimum constraints lifted (no obstacles for general methods) and less external knowledge required.
These virtues are crucial in building a flexible and generalized framework that effectively incorporates any existing design in augmentations. 

Inspired by such motivation, we thus propose a novel framework \textit{\textbf{M}eta \textbf{R}eweighting \textbf{Co}ntrastive ({\model})} model to \textit{(1)} reduce the noise from augmented samples by reweighting the augmentation loss; \textit{(2)} exploit the small weight augmented samples by contrasting the large and small weight augmented samples; and \textit{(3)} be accessible to any off-the-shelf augmentation methods and text representation learning methods.
As mentioned earlier, directly minimizing the reweighted loss will cause the reweight model collapse. It will take a shortcut by assigning the small weights for all the augmented samples. To overcome this issue, 
we consider the optimization of reweighted loss as the bilevel optimization, like recent meta reweighting works~\cite{shu2019meta, zheng2019meta, shu2020leveraging}. In the inner optimization loop, the main module, which is specific for the downstream task, is trained with the reweighted loss; while, in the outer optimization loop, the reweight module, also called meta reweighting module, assigns the proper weight for the augmented samples so the whole framework can achieve better performance on the held-out raw training dataset~(meta dataset). 
%
%
To fully exploit the augmented samples, we apply contrastive learning~\cite{he2020momentum} to refine augmented samples' representations. Specifically, we consider any low-weighted augmented samples within the same class as the negative samples (wider global restriction on negative samples), while the samples from the same originated samples as the positive samples (stricter local restriction on positive samples). This fine-grained contrastive learning allows the model to reinforce the difference between high-quality samples and low-quality samples.
Overall, {\model} differs from prior works in resolving the restrictions from many perspectives while still reaching high-level performance boosts: \textit{(1)}~It differs from macro-level augmentation noise reduction because it is capable of embracing as many augmented samples as possible; and \textit{(2)}~It also differs from the micro-level approach in terms of including all the augmented samples into model training.
%
In summary, the contributions of this work are four-fold:
\squishlist
\item To the best of our knowledge, we are the first to propose a framework combining both meta reweighting and contrastive learning in one unified deep learning model. Our framework leveraged bilevel optimization for automatically learning the weight of each augmented sample based on its quality. Then, our contrastive learning component leveraged the quality information to narrow down the gap between original data (golden, best quality) and the high-quality augmented data while enlarging the distance between the original data and low-quality augmented data.
\item To fully utilize the weight/quality information in the contrastive design, we proposed a novel weight-dependent dequeue-enqueue algorithm called \textit{\textbf{L}ifetime \textbf{A}ware \textbf{S}mallest \textbf{W}eight (\textbf{LASW})} algorithm. We elaborate the details of our design in Section~\ref{sec: framework}.
\item Our framework design lifted the minimum requirements to its surrounding components and generalized well: (1) It expects no specific data distribution (no distribution function leveraged); (2) It treats augmentation methods as black boxes; (3) It embraces as many augmentation methods and augmented samples as possible; and (4) It requires minimum adaptation efforts as a plug-in to any off-the-shelf classification models (i.e., it is highly cooperative with existing designs/models).
\item  We conducted extensive experiments on seven GLUE benchmark datasets. As a result of the advantages in the design mentioned above, our framework achieved an average of 1.6\%, up to 4.3\% absolute improvement on Text-CNN encoders, and an average of 1.4\%, up to 4.4\% absolute improvement on RoBERTa-base encoders on the benchmark datasets compared with the best baseline. A detailed analysis also provided more insights. 
\squishend
\begin{figure*}[!th]
	\centering
	\includegraphics[width=.7\linewidth]{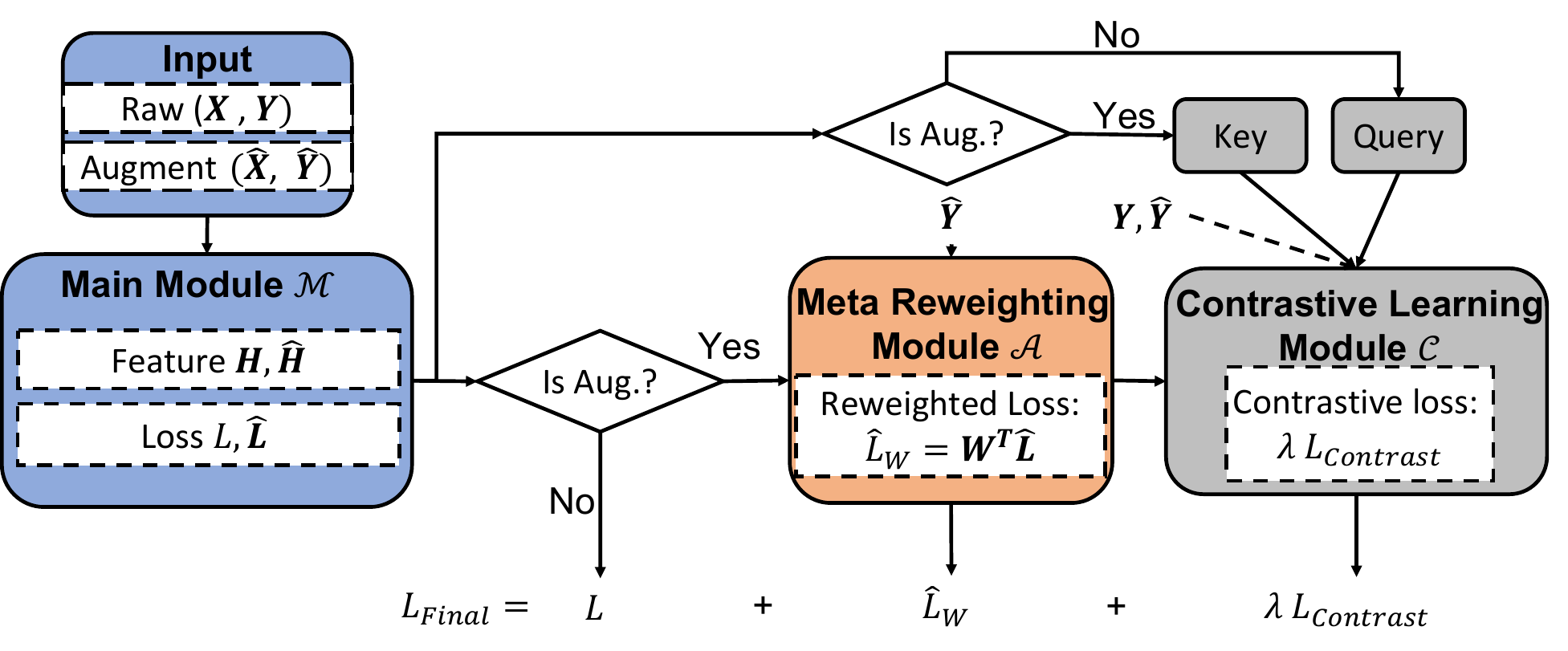}
	\vspace{-5pt}
	\caption{A high-level view of the overall framework. }
	\label{fig: main}
	\vspace{-5pt}
\end{figure*}
We organize the rest of this paper in the following sections. We formulate common terminologies in Section~\ref{sec: terminology}. We then introduce our designs in Section~\ref{sec: framework}. Experiment results are described in Section~\ref{sec: experiment}. Further analysis is presented in Section~\ref{sec: analysis}. We describe related work in Section~\ref{sec: relatedWork}. Lastly, we conclude our research in Section~\ref{sec: conclusion}.

\section{Notation Terminology}
\label{sec: terminology}
In this paper, we utilize the bold uppercase letters to denote the vectors or matrices (e.g., $\mathbf{Y}$), chancery fonts to represent the sets (e.g., $\mathpzc{{D}}$) and calligraphic fonts to denote the module (e.g., $\mathcal{M}$). For other notations, we will illustrate them in the relevant sections.  Formally, given the raw/original samples $\mathpzc{D}=\{(x_i, y_i)\}_{i=1}^{|\mathpzc{D}|}$ and augmented samples $\hat{\mathpzc{D}}=\{(\hat{x}_j, \hat{y}_j)\}_{j=1}^{|\hat{\mathpzc{D}}|}$, we train the classification model $\mathcal{M}$. The $x/\hat{x}$ is an instance of text and $y/\hat{y}$ is the class label. $({\textbf{X}}, {\textbf{Y}})$ and $(\hat{\textbf{X}}, \hat{\textbf{Y}})$ are the mini-batch for optimization.
For the raw dataset $\mathbf{\mathpzc{D}}$ and augmented dataset $\hat{\mathpzc{D}}$, there are $|\mathbf{\mathpzc{Y}}|$ different classes.


\section{Framework}
\label{sec: framework}
Given original/raw input text and their task-specific label $(\mathbf{X, Y})$ in benchmark datasets, augmentation methods will generate augmented samples represented as $(\mathbf{\hat{X}, \hat{Y}})$. While we generally believe the raw inputs are of golden quality, those augmented samples often contain noise. Our framework will consume both raw inputs and the augmented samples and automatically minimize the negative impact of noise within augmented samples. Thus, eventually, such a design will effectively boost the prediction/classification performance concerning the given task.
A high-level view of our overall framework is shown in Fig.~\ref{fig: main}. It mainly contains 3 parts:
\squishlist
\item \textit{Main module:} it learns the text representation $\mathbf{h}_i$ of the input $x_i$ and make the prediction $p(y_i)$ for $x_i$ and the main module can be any off-the-shelf text encoder; 
\item \textit{Meta reweighting module:} to reduce the noise brought by low-quality augmented samples, it assigns a weight for each augmented sample's loss, where the high-quality augmented samples will receive large weights while low-quality augmented samples will receive small weights.
\item \textit{Contrastive learning module:} it utilizes contrastive learning to enlarge the distance between high-quality augmented samples and low-quality augmented samples on weights. This helps the model fully exploit the low-weight augmented samples, which the meta reweighting learning model will tend to ignore.
\squishend

\smallskip\noindent We elaborate on the last two modules in the following sections.
\subsection{Meta Reweighting Module}
\noindent\textbf{Reweight-Plugin:} Researchers often assumed that the outcome of data augmentations was class-invariant. However, this assumption does not always hold, and simply treating all the augmented samples with the same importance is sub-optimal~\cite{weizou2019eda}. In addition, in some cases, although the augmented samples are consistent with the label, they introduce unrecognized noise as a side effect. The noise level within the augmented samples is not uniformly distributed, and it hurts the model's final performance.

Inspired by the previous work~\cite{zheng2021meta, shu2020leveraging}, to properly exploit these augmented samples, we propose a new reweight module $\Alpha$, assigning the importance weight $\mathbf{W}$ to these augmented samples. The reweight module will be only applied to augmented samples $\hat{\mathpzc{D}}$ during training, while it will be discarded for the raw samples $\mathpzc{D}$ or during evaluation. Concretely, given the augmented samples' hidden representations $\hat{\mathbf{H}}$ and labels for these samples $\hat{\mathbf{Y}}$, the reweight process is formed as:
\begin{equation}
\small
	\begin{aligned}
		& \mathbf{W} = \Alpha(\hat{\mathbf{H}}, \hat{\mathbf{Y}}) \\
		& \Alpha \rightarrow \sigma\left(\operatorname{MLP}([\hat{\mathbf{H}};\mathcal{E}(\hat{\mathbf{Y}})])\right)
	\end{aligned}
	\label{eq: reweight}
\end{equation}
where $\sigma$ is the $sigmoid$ activation function, $[\cdot\,;\,\cdot]$ is vector concatenation, $MLP$ is a multi-layer perceptron, and $\mathcal{E}$ is the embedding layer of the label, mapping each label into a dense vector $\hat{\mathbf{h}}_{label} \in \mathbb{R}^d$. In practice, we found that, for $MLP$, a shallow network structure ($\#layer < 3$) with dropouts is sufficient in reaching satisfying performance. The additional label embedding $\mathcal{E}$ allows us first learn the global representation of each class, and then allows $MLP$ to capture the interaction between augmented samples' hidden representation $\hat{\mathbf{H}}$ and given class $\hat{\textbf{Y}}$. 


\begin{figure}
	\centering
	\includegraphics[width=.9\linewidth]{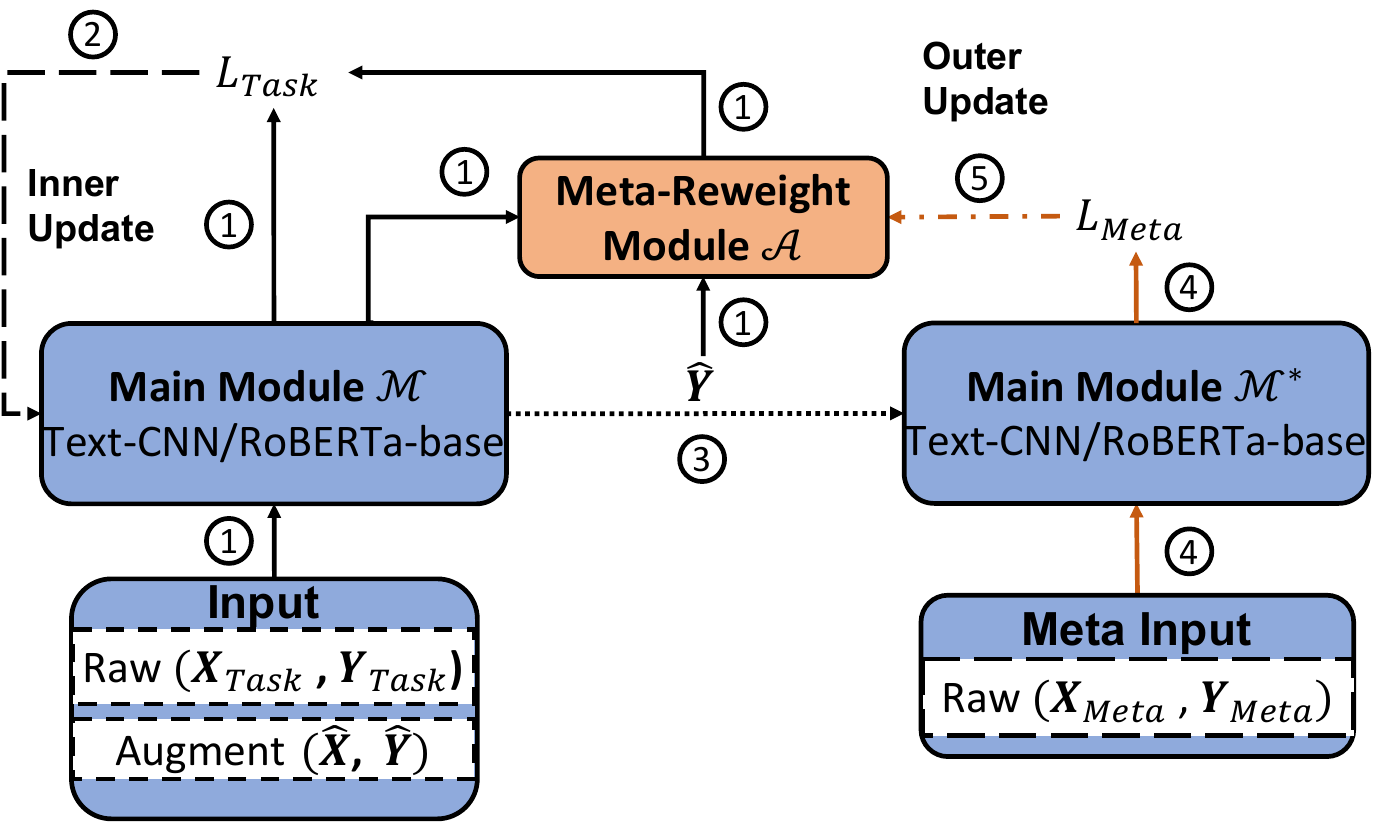}
	\caption{Approximated bilevel optimization procedure for meta reweighting module $\Alpha$: \textcircled{\raisebox{-0.9pt}{1}} utilize raw instances $(\mathbf{X}_{Task}, \mathbf{Y}_{Task})$ and augmented instances $(\hat{\mathbf{X}}, \hat{\mathbf{Y}})$ to do the forward pass for $L_{Task}$; \textcircled{\raisebox{-0.9pt}{2}} backward pass for $L_{Task}$ on $\mathcal{M}$, retaining the backward propagation graph for $\Alpha$; \textcircled{\raisebox{-0.9pt}{3}} update main module $\mathcal{M}$ to $\mathcal{M}^*$ through backward propagation; \textcircled{\raisebox{-0.9pt}{4}} utilize the raw meta input $(\mathbf{X}_{Meta}, \mathbf{Y}_{Meta})$ forward pass for $L_{Meta}$; \textcircled{\raisebox{-0.9pt}{5}} backward pass for $L_{Meta}$ to update meta reweighting module $\Alpha$.}
	\label{fig:meta}
	\vspace{-5pt}
\end{figure}

\smallskip\noindent\textbf{Objective Functions:} 
The objective function of classification is the linear combination of raw samples' loss $L$ and the reweighed augmentation loss $\hat{L}_W$  : 
\begin{equation}
\small
	\begin{aligned}
		& {L}_{Task}  = {L} + \hat{L}_{W}  \\
		& = {L} + \mathbf{W}^T\hat{\mathbf{L}}  \\
		& = \mathbb{E}_{({x}, {y})\in \mathcal{D}} l(y, p(y)) + \mathbb{E}_{(\hat{x}, \hat{y}) \in \hat{\mathcal{D}}, w \in \mathbf{W}}\, w\, l(\hat{y}, p(\hat{y}))
		\label{eq:reweight_loss}
	\end{aligned}
\end{equation}
where $l$ is the cross-entropy loss and $\hat{\textbf{L}}$ is the loss vector for all the augmented samples without reduction.
Directly minimizing Eq.~\ref{eq:reweight_loss} will cause the model to collapse as the reweight plugin $\Alpha$ will assign small (even zero) weights for all the augmented samples to reduce $\hat{L}_{w}$ without meaningful learning, and the main module will only learn from the raw samples since the gradients from the augmented samples are negligible.

\smallskip\noindent\textbf{Meta-Optimization:} To prevent the collapse problem of reweight module $\Alpha$ during optimization, we follow recent work on sample loss reweighting~\cite{liu2019darts, hospedales2020metalearning}, formulating this problem as a bilevel optimization problem. The optimal reweight module $\Alpha$ should achieve good performance on unseen raw samples.  To evaluate the effectiveness of the reweight module, we hold out a subset $\mathpzc{D}_{Meta} \subseteq \mathpzc{D}$.
After main module $\mathcal{M}$ and reweight module $\Alpha$ trained on $\mathpzc{D}_{Task}=\mathpzc{D}-\mathpzc{D}_{Meta}$ and relevant augmented samples $\hat{\mathpzc{D}}_{Task}$, 
the main module $\mathcal{M}$ should achieve better performance on $\mathpzc{D}_{Meta}$.
Thus, the objective function of bilevel optimization is formulated as:
\begin{equation}
\small
	\begin{aligned}
		\min _{\theta_\Alpha} & L_{Meta}\left(\theta_\mathcal{M}^{*}(\theta_\Alpha), \theta_\Alpha, \mathpzc{D}_{Meta}\right) \\
		\text { s.t. } & \theta_\mathcal{M}^{*}(\theta_\Alpha)=\operatorname{argmin}_{\theta_\mathcal{M}} L_{Task}(\theta_\mathcal{M}, \theta_\Alpha, \mathpzc{D}_{Task}, \hat{\mathpzc{D}_{Task}})
		\label{eq:bilevel}
	\end{aligned}
\end{equation}
where $\theta_{[\cdot]}$ is the parameters of the relevant module.
The outer loop update for reweight module $\theta_{\Alpha}$ will require the optimal main module $\theta^*_{\mathcal{M}}$ at every iteration, which is computationally infeasible. To reduce the computation requirement, we approximate optimal $\theta^*_{\mathcal{M}}$ by  one-step stochastic gradient descent~(SGD):
\begin{equation}
	\theta^*_{\mathcal{M}}(\theta_{\Alpha}) \approx \theta_{\mathcal{M}} - \alpha \, \nabla_{\theta_{\mathcal{M}}} L_{Task}(\theta_\mathcal{M}, \theta_\Alpha, \mathpzc{D}_{Task}, \hat{\mathpzc{D}})
\end{equation}
where $\alpha$ is the meta learning rate.
The approximated bilevel optimization procedure is illustrated in Fig.~\ref{fig:meta},
the inner optimization (steps \textcircled{\raisebox{-0.9pt}{1}} to \textcircled{\raisebox{-0.9pt}{3}}) is task-dependent, optimizing the parameters $\theta_\mathcal{M}$ by one-step SGD.
The outer optimization (steps \textcircled{\raisebox{-0.9pt}{4}} and  \textcircled{\raisebox{-0.9pt}{5}}) tries to update the meta reweighting module $\Alpha$ parameters $\theta_\mathcal{A}$ based on the updated main module $\theta^*_\mathcal{M}$ so the inner learning algorithm can better fit outer objective $L_{Meta}$.
%



\begin{figure}[t]
	\centering
	\includegraphics[width=.9\linewidth]{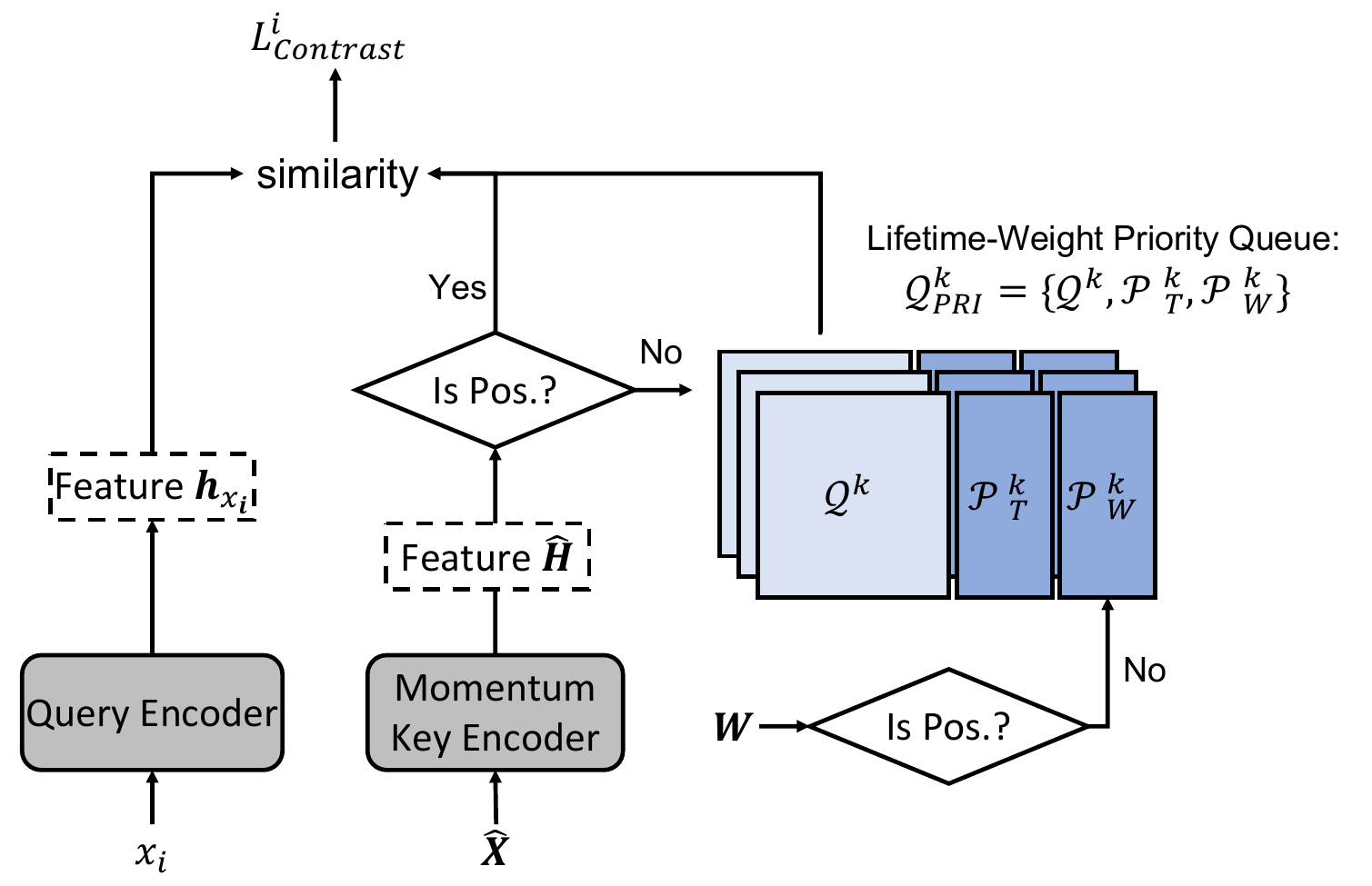}    \caption{The structure of contrastive learning module. }
	\label{fig:contrastive_learning}
	\vspace{-5pt}
\end{figure}
\subsection{Contrastive Learning Module}
\label{sec: contrastive}
Theoretically, the optimal {meta reweighting module} $\Alpha$ will assign small weights for low-quality samples. Although this allows the {main module} $\mathcal{M}$ to ignore the influence of the noise brought by these low-quality samples (negligible gradients in the model's back-propagation), it also neglects the potential contribution of the noised samples. Inspired by the recent work in contrastive learning~\cite{he2020momentum, robinson2021contrastive}, contrasting the distance among the query instance, negative instances, and positive instances to learn better feature representation, we believe the small weights can be emphasized by contrastive learning. Generally, we consider the small weighted augmented samples as negative instances while large weighted ones as positive instances. By contrasting the large weighted and small weighted augmented samples, these small weighted augmented samples can further contribute to better feature representation learning.

\begin{algorithm*}[t]{
\small
\SetKwInOut{Input}{Input}
\SetKwInOut{Output}{Output}
\SetAlgoLined
\caption{LASW Dequeue-Enqueue Algorithm}
\Input{Weight-lifetime priority queues $\mathcal{Q}_{PRI} = \{\mathcal{Q}^k_{PRI}\}_{k=1}^{|\mathpzc{Y}|}$, where $\mathcal{Q}^k_{PRI}=(\mathcal{Q}^k, \mathcal{P}^k_W, \mathcal{P}^k_T)$, 
maximum lifetime $\tau$,  augmented mini-batch $\{\hat{\mathbf{H}}, \hat{\textbf{Y}}\}$ and weight $\mathbf{W}$. }
\Output{Updated $\mathcal{Q}_{PRI}$.}
\begin{algorithmic}[.9]
\FOR{$k$ in $\mathpzc{Y}$}
    \STATE Choose $\hat{\mathbf{H}}^k$ and $\mathbf{W}^k$, where $\hat{\mathbf{Y}}==k$; \hfill\COMMENT{Select enqueue candidates}
    \STATE $\mathcal{P}_T^k = \mathcal{P}_T^k - 1$; \hfill\COMMENT{Update instance's lifetime}
    \STATE Dequeue $N$ instances from  $\mathcal{Q}^{k}$, where $\mathcal{P}_T^k <= 0$ and discard them;\tikzmark{top1} \hfill\COMMENT{\textbf{Lifetime aware dequeue-enque}}
    \STATE Enqueue $Top-N$ smallest weight instances $\hat{\mathbf{H}}_N^k$ from $\hat{\mathbf{H}}^k$ by $\mathbf{W}^k$;\tikzmark{right1}\tikzmark{bottom1} 
    \FOR{$\hat{\mathbf{h}}^k$ in $\hat{\mathbf{H}}^k_{\lnot_{N}}$} 
        \IF{len($\mathcal{Q}^{k}{[\mathcal{P}_W^k > \mathbf{W}^k_{\lnot_{N}}]}$) $> 0$}
        \STATE Dequeue the largest weighted instance from $\mathcal{Q}^{k}{[\mathcal{P}_W^k > \mathbf{W}^k_{\lnot_{N}}]}$ and discard it;  
        \hfill\COMMENT{\textbf{Small weight dequeue-enque}}
        \STATE Enqueue $\hat{\mathbf{h}}^k$;
        \ENDIF
        \STATE Set $\mathcal{P}_T^k=\tau$ and  $\mathcal{P}_W^k=\mathbf{W}^k_{\lnot_{N}}$ for  newly enqueued samples from $\hat{\mathbf{H}}^k_{\lnot_{N}}$;
    \ENDFOR\tikzmark{bottom2}

    \ENDFOR
\end{algorithmic}
 \label{alg: opt}
 }
\end{algorithm*}
\vspace{-5pt}

\subsubsection{Prior Knowledge of Contrastive Learning}
Given a query sample $x_i$, positive augmented samples ${\hat{\textbf{X}}}_{Pos}^i$, and negative augmented samples ${\hat{\textbf{X}}_{Neg}}$, the overall objective function of contrastive learning is:
%
\begin{equation}
\begin{aligned}
    {L}_{Contrast} = &
    -\mathbb{E}\left[\frac{1}{|\hat{\textbf{X}}_{Pos}^i|}\sum_{\hat{x}_{p}^i \in \hat{\textbf{X}}_{Pos}^i} \log {P}_{Contrast}\right] \\
    {P}_{Contrast} =  & \frac{\exp \left(\mathbf{h}_{x_i}^T\mathbf{h}_{\hat{x}_p^i}\right)}{ \sum_{\hat{x}\in \hat{\textbf{X}}_{Neg}\bigcup\{\hat{x}_p\}} \exp \left(\mathbf{h}_{x_{i}}^T\mathbf{h}_{\hat{x}}\right)}
\end{aligned}
\end{equation}
Negative samples are not instance dependent, as $\hat{\mathbf{X}}_{Neg}$ are usually stored into a large queue~\cite{he2020momentum} and can be shared across instances, while the positive samples  $\hat{\mathbf{X}}^i_{Pos}$ are usually originated from query sample $x_i$. 
To achieve effective performance, the existing approaches~\cite{he2020momentum, qu2020coda} often include a large queue $\mathcal{Q}$ 
to store the historical hidden representation of negative augmented samples $\hat{\mathbf{H}}_{Neg}$, and a Siamese network design to encode the query sample $x_i$ and augmented/key samples $\hat{\mathbf{X}}\supseteq \hat{\mathbf{X}}_{Pos}^i + \hat{\mathbf{X}}_{Neg}$.
Concretely, let $\theta_\mathcal{M}$ and $\hat{\theta}_\mathcal{M}$ be the parameters of query encoder (i.e., the main module) and key encoder, respectively. We update the key encoder's parameters with momentum rule:
\begin{equation}
	\hat{\theta}_\mathcal{M} =\gamma{\hat{\theta}_\mathcal{M}} + (1-\gamma){\theta}_\mathcal{M}
\end{equation}
where $\gamma$ is the ratio controlling the update step size.

\subsubsection{Weight dependent Contrastive Learning} 
~\\
\noindent\textbf{Difference from prior works:} Existing works take augmented samples originated from different raw samples~\cite{he2020momentum} or different classes~\cite{khosla2020supervised} as negative samples. Our design, however, consider small weighted augmented samples within the same class as negative samples. The purpose of requiring ``within the same class'' is two-fold. First, it is infeasible to compare the weight across classes. For example, in sentiment classifications, a small weighted positive augmented sample may have less contribution on meta-dataset ${\mathpzc{D}}_{Meta}$. However, it is not necessarily close to the negative class. Second, the class difference may dominate the weight difference~\cite{khosla2020supervised}, and the contrastive learning module will only capture the coarse class difference however ignore the fine-grained weight difference. 
What is more, existing works are limited by their simple queue structure and FIFO dequeue-enqueue operations. We, however, designed lifetime-weight priority queues for different classes as discussed below.

\smallskip\noindent\textbf{Our design:} We show our new contrastive learning in Fig.~\ref{fig:contrastive_learning}.
Concretely, given a query sample $x_i$ in class $k$: (1) we propose the small weighted augmented samples also in class $k$ as the negative augmented samples $\hat{\mathbf{X}}_{Neg}$. They will be stored in the negative queue $\mathcal{Q}_{PRI}^k$; and (2) following~\cite{he2020momentum, qu2020coda}, we consider the large weighted augmented samples originated from $x_i$ as positive augmented samples $\hat{\mathbf{X}}_{Pos}^i$.
It is worth noting that $x_i$, $\hat{\textbf{X}}_{Pos}^i$ and $\hat{\textbf{X}}_{Neg}$ are from the same class $k$. 

Our new approach allows us to capture the implicit knowledge from the noisy samples, and provides a fine-grained way to learn better feature representation inside each class. We designed lifetime-weight priority queues for $|\mathpzc{Y}|$ different classes $\mathcal{Q}_{PRI}=\{\mathcal{Q}^k_{PRI}\}_{k=1}^{|\mathpzc{Y}|}$. Specifically, for class $k$, the priority queue is $\mathcal{Q}^k_{PRI} = (\mathcal{Q}^y, \mathcal{P}_{T}^k, \mathcal{P}_{W}^k)$, where $\mathcal{Q}^y$ will store the negative samples, $\mathcal{P}_W^k$ is the lifetime priority index and $\mathcal{P}_{W}^k$ is the weight priority index. 
In addition, we also design a Lifetime Aware Small Weight (LASW) dequeue-enqueue algorithm to properly keep a queue of up-to-date small weight negative instances. 

\begin{table*}[t]
	\centering
	\small
	\caption{Important hyperparameters used in our contrastive learning module.}
	\scalebox{.9}{
	\begin{tabular}{l | c | l} \hline
		Name & Notation & Brief Description \\ \hline
		Positive Ratio & $\rho$ & The proportion of high-weight augmented samples treated as positive samples. \\
		Queue Size & $N_{Neg}$ & The queue size of negative (low-weight) samples for each class. \\
		Survive-Time-slot & $\tau$ & The maximum time for an augmented sample to live in a queue. \\
		Importance Score & $\lambda$ & The weight of contrastive loss $L_{Contrast}$ in the overall loss function Eq.~\ref{eq: overall}.\\ \hline
	\end{tabular}}
	\label{tab:hyper}
\end{table*}

\begin{table*}[!ht]
	\centering
	\small
	\caption{Dataset Statistics.}
	\scalebox{.95}{
	\begin{tabular}{l|c|c|c|c|c|c|c} \hline
		Task & RTE & MRPC & CoLA & SST-2 & QNLI & QQP & MNLI-m \\ \hline
		Train & 2.5k & 3.7k & 8.6k & 67.4k & 105k & 364k & 393k \\
		Aug. & 13.7k & 22.0k & 68.4k & 533k & 627k & 2.18m & 2.34m \\
		Aug. Ratio & 5.50 & 6.00 & 8.00 & 7.91 & 5.98 & 5.97 & 6.00 \\ \hline
		Dev & 278 & 408 & 1.1k & 873 & 5.5k & 40k & 9.8k \\ \hline
	\end{tabular}}
	\label{tab:GLUE dataset}
	\vspace{-5pt}
\end{table*}
\smallskip\noindent\textbf{LASW dequeue-enqueue Algorithm.}
The LASW algorithm is presented in Algorithm~\ref{alg: opt}. It contains two pairs of dequeue-enqueue operations. 
For each class $k$, it will firstly do the \textit{Life time aware dequeue-enqueue} in step 4 and 5. It will dequeue instances from $\mathcal{Q}_{PRI}^k$ when instances' lifetime $\mathcal{P}_{T}^k$ less than 0. After the dequeued process, it will enqueue the same amount of  samples from the mini-batch in weight ascending order. This can help us keep the instances inside the queue up-to-date and with small weight. The second dequeue-enqueue operation is named as \textit{small weight dequeue-enqueue}, from step 6 to 12. It will compare the not enqueued instances $\hat{\mathbf{H}}^k_{\lnot_{N}}$ with all the instances inside $\mathcal{Q}^{k}$ on weight $\mathcal{W}^k_{\lnot_{N}}$,  dequeue the largest weighted instances from $\mathcal{Q}^{k}{[\mathcal{P}_W^k > \mathbf{W}^k_{\lnot_{N}}]}$ and enqueue these instances from the mini-batch.
After every enqueue operation, the module will reset the lifetime priority to $\tau$ and set the weight priority from $\mathbf{W}_k$ for these enqueued instances.



\smallskip\noindent\textbf{Positive sample selection.} Prior works generally consider all relevant augmented samples as valid positive samples of the raw sample. Such treatment can bring up problems with low-quality augmented samples, where the noise within these augmented samples can overwhelm the contributions or even cause label-flipping problems. We set up an extra selection ratio $\rho \in [0,1]$ as a hyperparameter, which will choose the top $\rho$ augmented samples from the mini-batch as positive samples concerning their calculated weights.

\subsection{Overall Objective Function}
To learn a better feature representation for the downstream tasks, we consider integrating the classification loss and contrastive loss and optimize them together. We can define the final objective function as:
\begin{equation}
	{L}_{Final} = {L}_{Task} + \lambda \, {L}_{Contrast}
	\label{eq: overall}
\end{equation}
where $\lambda$ is the hyper-parameter to control the importance of contrastive learning in the tasks. If $\lambda=0$, this means the model only reduces the noise from augmented samples by reweighting these augmented samples. It should be noticed that doing the backpropagation on $L_{Final}$ will only update the main module $\mathcal{M}$ to avoid reweight module $\Alpha$ collapse. We update $\Alpha$ by Eq.~\ref{eq:bilevel} instead of Eq.~\ref{eq:reweight_loss}.

\smallskip\noindent\textbf{Important Hyperparameters.}
To further help readers better understand our novel design in the contrastive learning module, we summarize our newly introduced/invented hyperparameters in Table~\ref{tab:hyper} with their names, notations, and brief descriptions. We will also conduct a deeper analysis of these hyperparameters in Section~\ref{sec: analysis} to reveal their non-trivial contributions as well as the impact of varying their values in terms of overall performance.

\begin{table*}[t]
	\centering
	\small
	\caption{Experiment results on GLUE benchmark dataset in format $avg_{std}$. Best results: \textbf{BOLD}. Second-best: \underline{UNDERLINED}. Overall result is the row average of all tasks. All results rounded to precision of 0.001.}
	\scalebox{.9}{
	\begin{tabular}{l|c|c|c|c|c|c|c|c} \hline
		{Methods} &  RTE(ACC) & MRPC(ACC) & CoLA(MCC) & SST-2(ACC) & QNLI(ACC) & QQP(ACC) & MNLI-m(ACC) & Overall
		\\ \hline \hline
		Text-CNN & $.515_{.010}$ & $.702_{.006}$ & $.187_{.013}$ & \underline{$.837_{.003}$} & $.604_{.005}$ & \underline{$.791_{.001}$} & $.542_{.001}$ & .598 \\ \hline
		+ Aug. & $.524_{.002}$ & $.698_{.005}$ & $.194_{.011}$ & $.830_{.002}$ & $.600_{.001}$ & $.794_{.002}$ & $.528_{.001}$ & .594 \\
		+ Aug. \& Filter & $\underline{.527}_{.011}$ & $\underline{.724}_{.000}$ & $\underline{.202}_{.004}$ & $\underline{.855}_{.001}$ & $\underline{.615}_{.001}$ & $\underline{.796}_{.001}$ & $\underline{.551}_{.001}$ & \underline{.610} \\
		\model & $\textbf{.534}_{.001}$ & $\textbf{.735}_{.002}$ & $\textbf{.213}_{.001}$ & $\textbf{.866}_{.001}$ & $\textbf{.632}_{.001}$ & $\textbf{.839}_{.002}$ & $\textbf{.565}_{.001}$ & \textbf{.626} \\ \hline \hline
		RoBERTa-base & $.713_{.023}$ & $.882_{.005}$ & $.566_{.013}$ & $.944_{.003}$ & $.924_{.007}$ & $.912_{.001}$ & $.875_{.001}$ & .830 \\ \hline
		+ Aug. & $.742_{.001}$ & $.880_{.004}$ & $.563_{.001}$ & $.953_{.002}$ & $.921_{.001}$ & $\textbf{.915}_{.001}$ & $\underline{.876}_{.002}$ & .836\\
		+ Aug. \& Filter & $\underline{.744}_{.002}$ & $\underline{.883}_{.004}$ & $.583_{.007}$ & $\underline{.956}_{.001}$ & $\underline{.929}_{.001}$ & $\textbf{\underline{.915}}_{.001}$ & $.875_{.002}$ & \underline{.845}\\
		CERT (origin) &  $.722_{.--}$ & $--$ &  $\underline{.629}_{.--}$ & $.936_{.--}$ & $.923_{.--}$ & $.914_{.--}$ & $.863_{.--}$ & $--$ \\
		SCL (origin) & $--$ & $--$ & $--$ & $.949_{.006}$ & $.925_{.004}$ & $--$ & $.853_{.005}$ & $--$\\
		\model & $\textbf{.760}_{.004}$ & $\textbf{.902}_{.009}$ & $\textbf{.673}_{.002}$ & $\textbf{.960}_{.001}$ & {$\textbf{.930}_{.002}$} & {$.894_{.001}$} & {$\textbf{.893}_{.001}$} & \textbf{.859}\\
		\hline
	\end{tabular}}
	\label{tab:GLUE result}
	\vspace{-5pt}
\end{table*}

\section{Experiment}
\label{sec: experiment}
With all the above descriptions and explanations of our network design, we can now verify the effectiveness of our framework {\model}. We conduct experiments on tasks from General Language Understanding Evaluation (GLUE) benchmark datasets~\cite{wang-etal-2018-glue}. We compare {\model} with the state-of-the-art approaches on learning from noised augmented data.

\subsection{Datasets and Setup}
\noindent\textbf{Datasets.} The natural language understanding benchmark dataset GLUE~\cite{wang-etal-2018-glue} contains 9 tasks/sub-datasets, ranging from single-sentence tasks to similarity and paraphrase tasks and inference tasks. Following CoDa~\cite{qu2020coda}'s setting, we only focus on 7 out of the 9 tasks: CoLA, SST-2, MRPC, QQP, MNLI-m, QNLI, and RTE. Note that WNLI was not considered in CoDa, and STS-B is a regression task that is not the scope of this paper. We train the models on the given training set and report the results on the given development set. Evaluation metrics are predefined/given by each task. A brief statistics description of these tasks is shown in Table~\ref{tab:GLUE dataset}. The amount of augmented data depends on the length of each raw sample and the task properties. For example, a raw instance with a longer text string will be easier to augment and thus have more augmentation variations than shorter raw instances.

\smallskip\noindent\textbf{Setup.} We utilize the Text-CNN~\cite{kim2014convolutional} and RoBERTa-base~\cite{liu2019roberta} as our backbone/encoder main module for text feature extraction and classification. However, {\model} can easily be applied to other text encoders. We use Adam as our optimizer.
For GLUE benchmark datasets, we utilize 5 representative data augmentation methods to augment the raw data, namely:
\squishlist
\item \textit{Wordnet}~\cite{miller1995wordnet}: a method using Wordnet as a dictionary of synonyms to replace candidate words.
\item \textit{Easydata}~\cite{weizou2019eda}: a method using the combination of word synonym replacement, random word insertion, random word swap, and random word deletion.
\item \textit{Checklist}~\cite{ribeiro2020beyond}: a method using dictionary-based named entity substitution.
\item \textit{Embedding}~\cite{alzantot2018generating}: a method using word embeddings to search similar words for candidate word substitutions.
\item \textit{Charswap}~\cite{li2019textbugger}: a method with character-level noise injection methods (insertion, swap, deletion, and replacement).
\squishend
One can easily extend our framework to other augmentation methods as well. As we stated earlier, one of the advantages of our framework is that, given any main module encoder and any augmented samples, {\model} shall be able to enhance the model performance further.

\subsection{Baseline Methods}
Since our framework is augmentation agnostic, we compare the baseline methods to understand which method best exploit the augmented samples. We include two branches of baseline methods: contrastive learning and vanilla approaches.

\smallskip\noindent\textbf{Contrastive Learning Approaches:} \textbf{CERT}~\cite{fang2020cert} used MoCo~\cite{he2020momentum} and the augmented data for contrastive learning in the model pretraining phase, and then finetuned the adjusted pretrained model on the downstream task. It utilized a queue to store large amounts of negative samples and momentum update to learn a consistent feature representation of the en-queued negative samples. It considered the augmented samples from the same raw samples as positive samples and augmentation from other ones as negative ones.
\textbf{SCL}~\cite{gunel2021supervised}, a recent work, leveraged supervised contrastive learning in the finetuning phase. It considers the augmented samples from different classes as the negative samples, while the positive samples are obtained from the inner-class augmented samples.


\smallskip\noindent\textbf{Vanilla Approaches:}
This part includes the more straightforward cases of the plain main module (Text-CNN and RoBERTa-base) and their performance with augmentations with (\textit{+ Aug \& Filter}) or without (\textit{+ Aug})  manual empirical filter mechanisms. More specifically, we used Flesch Reading Ease~\cite{flesch1979write} readability score as the filter metric. We set the upper limit and lower limit of the score as hyper-parameters. Augmented samples with scores between the two limits will be kept while the framework will omit others. We find the best-performing combination with a thorough grid search.


\subsection{Main Results}
\label{sec: results}
We show the detailed experiment results in Table~\ref{tab:GLUE result}, where we used the abbreviations ``ACC'' standing for accuracy and ``MCC'' standing for Matthews Correlation Coefficient. We run each experiment 5 times with different random seeds and report the average and the standard deviation in the format of $avg_{std}$. For each different network backbone/encoder (\textit{main module}), we mark the best performing average result in bold and the second-best underlined. All results are rounded to a precision of 0.001. Rows noted with \textit{(origin)} are results directly obtained from the referenced source papers, where some results were unavailable/not provided by the authors. In this case, we marked the blank space as ``$--$'' (i.e., the authors of CERT and SCL papers did not test their models over some GLUE tasks). GLUE tasks are arranged in the same order as Table~\ref{tab:GLUE dataset}, where their training data volumes grow from left to right. We also calculate the macro average across tasks in the last column except for CERT and SCL because it is not fair to calculate the overall average/performance based on only certain task results.
Through these experiment results, we found several interesting observations:

\smallskip\noindent\textbf{Overall result: }\model ~showed superior performance against all the baselines and variations in almost all tasks. Most of the improvements between the best and second-best results are nontrivial concerning the averages and standard deviations. The most significant improvement is {\model} with RoBERTa-base under CoLA (MCC), where we observed a 4.4\% absolute performance boost on average compared with the best baseline. The performance boost is especially larger when the size of the original training set is relatively small (tasks on the left of the table). On the RoBERTa-base encoder, the performance enhancement ranges from 1.6\% to 4.4\% on the top 3 smallest tasks, while for the remaining 4 tasks, the improvement ranges from 0 to 1.7\%. This meets our intuition as deep learning models generally perform better with more provided data.

\smallskip\noindent\textbf{\model ~across different backbones:} \model ~is effective on both backbone main module architectures, showing an improvement across tasks. {\model} made an absolute performance boost of 2.8\% on Text-CNN backbones and 3.9\% on RoBERTa-base backbones, indicating that our model is highly adaptive to a wide range of deep learning designs.

\begin{figure*}[th]
	\centering
	\begin{subfigure}[t]{0.35\textwidth}
		\centering
		\includegraphics[width=\linewidth]{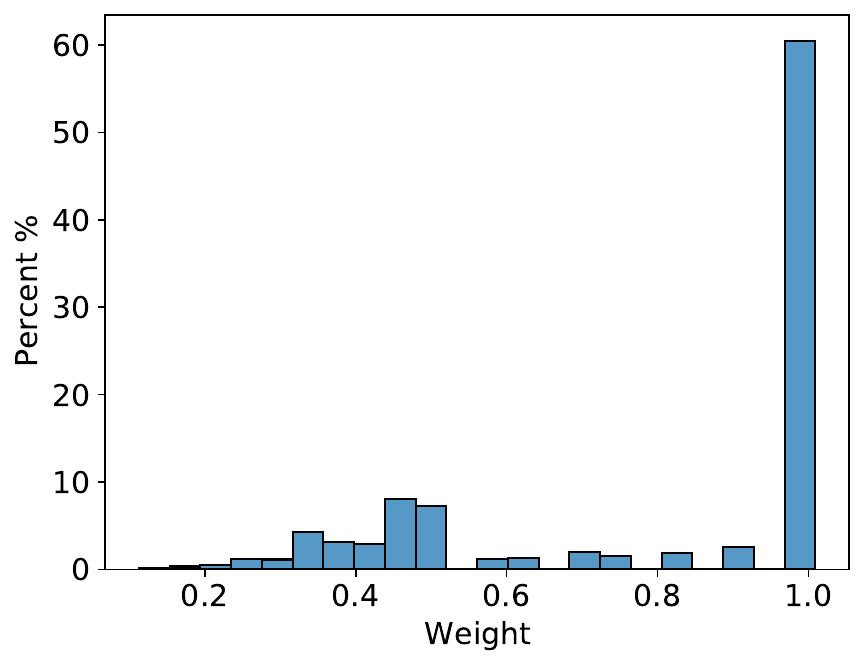}
		\caption{RTE}
	\end{subfigure}
	\begin{subfigure}[t]{0.35\textwidth}
		\centering
		\includegraphics[width=\linewidth]{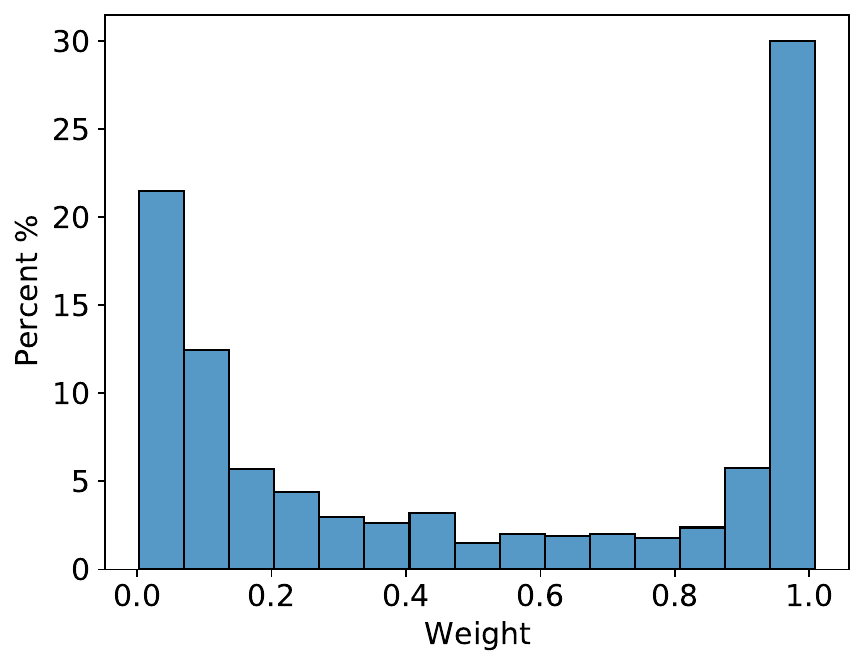}
		\caption{MRPC}
	\end{subfigure}
	\caption{Augmented samples weight distribution from meta reweighting module with basic encoder RoBERTa-base.}
	\label{fig:weight_analysis}
	\vspace{-5pt}
\end{figure*}

\smallskip\noindent\textbf{Plain modules vs. \textit{+ Aug.}:} It seems there are apparent differences in the capability of embracing augmented data. In general, RoBERTa-base methods are better at incorporating the augmented samples,   thus show overall performance improvement on the given metrics, while Text-CNNs struggle on several tasks. In particular, on SST-2, QQP, and QNLI, Text-CNN performed better without augmentations. These observations might point to the differing capability of models in tolerating noise within the data while still extracting useful information. Once again, as we observed before, the augmentations are less effective on tasks with larger training sets.

\smallskip\noindent\textbf{\textit{+ Aug.} vs. \textit{+ Aug. \& Filter}}: Generally speaking, augmentations without filters is actually a special case for applying filters where the limits are set to $(-\inf, \inf)$. Moreover, the original baseline without any augmentation can also be deemed a special case where the filtering rules are so strict that none of the augmented samples meet the requirement, returning an empty set on augmented samples. We found filters were helpful under most tasks. It is worth noting that on certain tasks, the \textit{+ Aug.} was not helpful, whereas \textit{+ Aug. \& Filter} managed to reverse the trend and boosted the performance (e.g., MRPC on both encoders and CoLA on RoBERTa-base).

\smallskip\noindent\textbf{{\model} vs. \textit{+ Aug \& Filter}:} Although \textit{+ Aug. \& Filter} boosted the performance in most tasks compared with the original backbones (i.e., Text-CNN and RoBERTa-base with or without augmentations), our model's reweighting strategy performed even better than it. 
The largest improvements over \textit{+ Aug. \& Filter} were 4.7\% on Text-CNN and 4.4\% on RoBERTa-base. The overall improvements are also strong, varying from 1.6\% on Text-CNN to 1.4\% on RoBERTa-base. One may easily observe that our {\model} dominated \textit{+ Aug. \& Filter} across most tasks with only two exceptions: with RoBERTa-base as the backbone, {\model} reached the same level of performance as \textit{+ Aug. \& Filter} on QNLI, as there is few statistical difference between the two models; and on QQP, {\model} slightly under-performed against \textit{+ Aug. \& Filter}. Such results reflected a similar conclusion as the prior works~\cite{fang2020cert, he2020momentum, gunel2021supervised}, where researchers found meta learning and contrastive learning tended to have more significant improvements on smaller datasets compared with larger datasets.

\smallskip\noindent\textbf{{\model} VS. existing approaches:} We tried to apply the given code provided by CERT~\cite{fang2020cert}, but we were not able to achieve satisfying results/the same results that the authors reported. Therefore, we only reported the original results published in CERT(origin). It is worth noting that the original paper leveraged a slightly different encoder BERT-base and a more advanced augmenting method (i.e., back-translation). Overall, the performance improvement of our framework over CERT is still apparent. A similar conclusion can be applied to SCL (origin), where \model ~also achieved better performance across tasks. We also tried to implement and run our version of other baselines such as MMEL~\cite{yi2021reweighting} and CoDa~\cite{qu2020coda} since they did not release their source code. Unfortunately, the models poorly performed compared with other baselines (e.g., CERT(origin) and SCL(origin)). Therefore, we do not report their results.

\begin{figure*}[t]
	\centering
	\begin{subfigure}[t]{.85\textwidth}
		\centering
		\includegraphics[width=\linewidth]{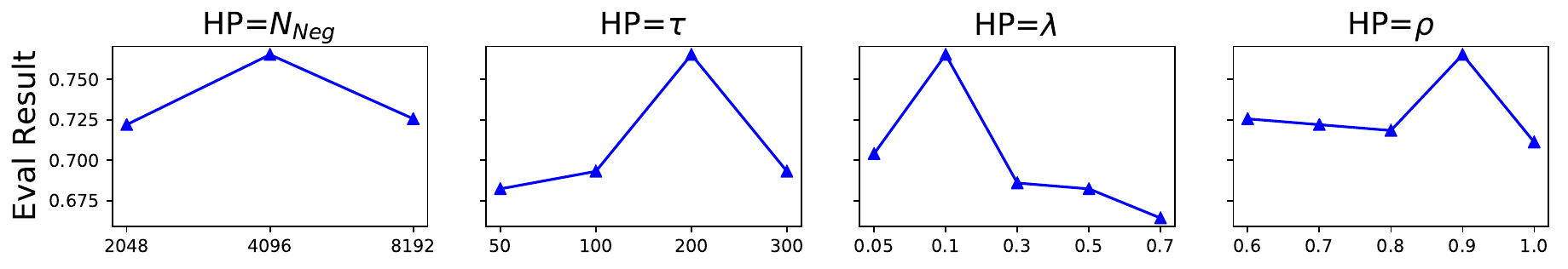}
		\caption{RTE}
		\label{fig:hyper_rte}
	\end{subfigure}\\
	\begin{subfigure}[t]{.85\textwidth}
		\centering
		\includegraphics[width=\linewidth]{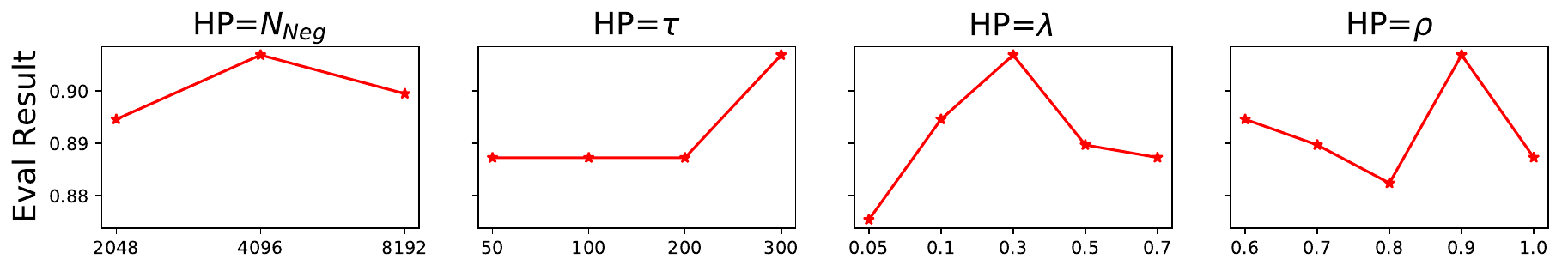}
		\caption{MRPC}
		\label{fig:hyper_mrpc}
	\end{subfigure}
	\caption{Hyperparameter~(HP) analysis of Contrastive Learning on RoBERTa-base encoder.}
	\label{fig:hyper_parameter}
	\vspace{-5pt}
\end{figure*}

\section{Analysis}
\label{sec: analysis}
In this section, we conduct extra experiments to shed more light on insights into our framework's success. We show a detailed analysis of each component's contribution in our framework and the training dynamics of \model.

\subsection{Meta Reweighting Module Analysis}
To understand whether the meta reweighting module can effectively capture the information within noisy data, we visualize the output weights of augmented samples generated from RTE and MRPC datasets. In Fig.~\ref{fig:weight_analysis}, we find that augmented samples of these two datasets have different weight distribution patterns. In RTE task, most of the augmented samples receive weights close to 1, while in MRPC task, many augmented samples receive weights less than $0.5$.
As we already showed in Table~\ref{tab:GLUE result} and stated in Section~\ref{sec: results}, in RTE task, the model with all the augmentations \textit{+Aug} achieved better performance than the model without augmentations. Such trend does not apply on MRPC, where augmentations \textit{+Aug} were not helpful. This indicates that the quality of MRPC's augmented data is not as good as RTE's one. Despite such a disadvantage, {\model} still improved its performance in both tasks by differentiating noisy samples via assigning them with lower weights. The weight distribution analysis confirmed the effectiveness of our meta reweighting module.

\subsection{Contrastive Learning Analysis}
\subsubsection{Hyperparameter Analysis}
\label{sec:parameter_analysis}
To better understand which parameters play an essential role in our contrastive learning module, we conduct several hyperparameter analysis on the number of negative samples $N_{Neg}$, the importance of contrastive learning loss $\lambda$, survive-time-slot $\tau$ and positive ratio $\rho$. Fig.~\ref{fig:hyper_parameter} shows how the evaluation result changed when we vary these hyperparameters under RoBRETa-base encoder. In the following paragraphs, we describe their impacts on experiment results by referring to column-wise subfigures in Fig.~\ref{fig:hyper_parameter}.

\smallskip\noindent\textbf{Impact of queue size $N_{Neg}$:} Existing works in contrastive learning~\cite{fang2020cert, qu2020coda, he2020momentum} have shown that a large queue size guarantees better performance of a downstream task. Usually, the larger the queue size is, the better the model's performance is. Prior works thus adopt the largest possible queue size based on the training set size. However, our experiment results showed that the optimal queue size might not always be the largest possible number concerning the specific task. In our model, the best performance was reached under $N_{Neg}=4,096$, rather than the larger 8,192 under both tasks. Such interesting differences might orient from our novel design of contrastive learning processes and algorithm. Another benefit of a smaller queue size is that our model has fewer computation costs. 

\smallskip\noindent\textbf{Impact of survive-time-slot $\tau$:} We observe that our model achieves the best performance under $\tau=200$ in RTE task, while it achieves the best performance under $\tau=300$ in MRPC task. Such difference might be because the size of {MRPC} dataset is larger than {RTE}, so the small weight instances in MPRC would be considered proper negative samples for a much longer time compared with augmented samples in {RTE}.

\smallskip\noindent\textbf{Impact of important score $\lambda$:} We observe that properly combining the contrastive learning objective function and downstream task objective function yields the best performance, while the optimal values are more task-specific rather than globally general.

\smallskip\noindent\textbf{Impact of positive ratio $\rho$:} We noticed that when $\rho=0.9$, {\model} achieved the best performance on most tasks. However, when $\rho=1.0$ (i.e., considered all the same origination augmented instance as positive samples), {\model} did not show the best performance. It indicates the importance of a higher standard for constructing positive samples (i.e., only keeping high-quality augmented samples as positive samples). It also reveals that the importance of using the hyper-parameter $\rho$.

\begin{table}[!tbh]
	\centering
	\small
	\caption{Ablation study of contrastive learning w/ RoBERTa-base encoder.}
	\scalebox{.9}{
	\begin{tabular}{l|c|c} \hline
		\textbf{Methods on RoBERTa-base} &  RTE & MRPC \\ \hline
		+ Aug. \& Filter & $.744_{.002}$ & $.883_{.004}$\\
		\hline
		{\model} & $.760_{.004}$ & $.902_{.009}$ \\
		\hline
		\hline
		{\model} \textit{w/o} Contrastive Learning & $.722_{.001}$ & $.899_{.001}$ \\
		\hline
		{\model} \textit{w/o} LASW &  $.747_{.025}$ & $.889_{.006}$
		\\
		\hline
	\end{tabular}}
	\label{tab:no_contrastive}
\end{table}

\begin{figure*}[t]
	\centering
	\begin{subfigure}[t]{0.4\textwidth}
		\centering
		\includegraphics[width=\linewidth]{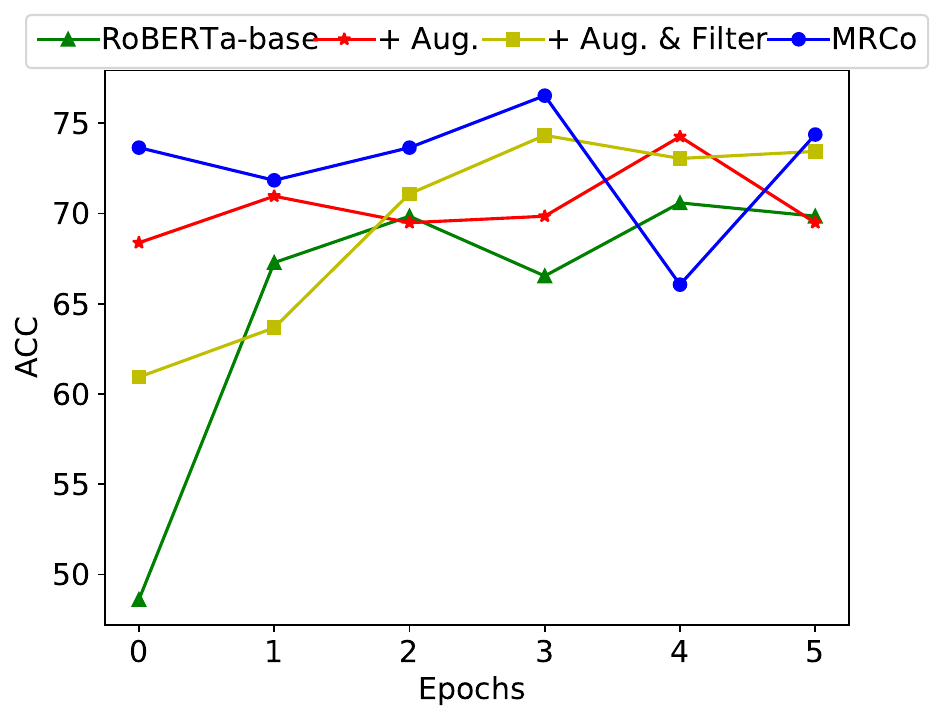}
		\caption{RTE}
		\label{fig:supervised}
	\end{subfigure}
	\begin{subfigure}[t]{0.4\textwidth}
		\centering
		\includegraphics[width=\linewidth]{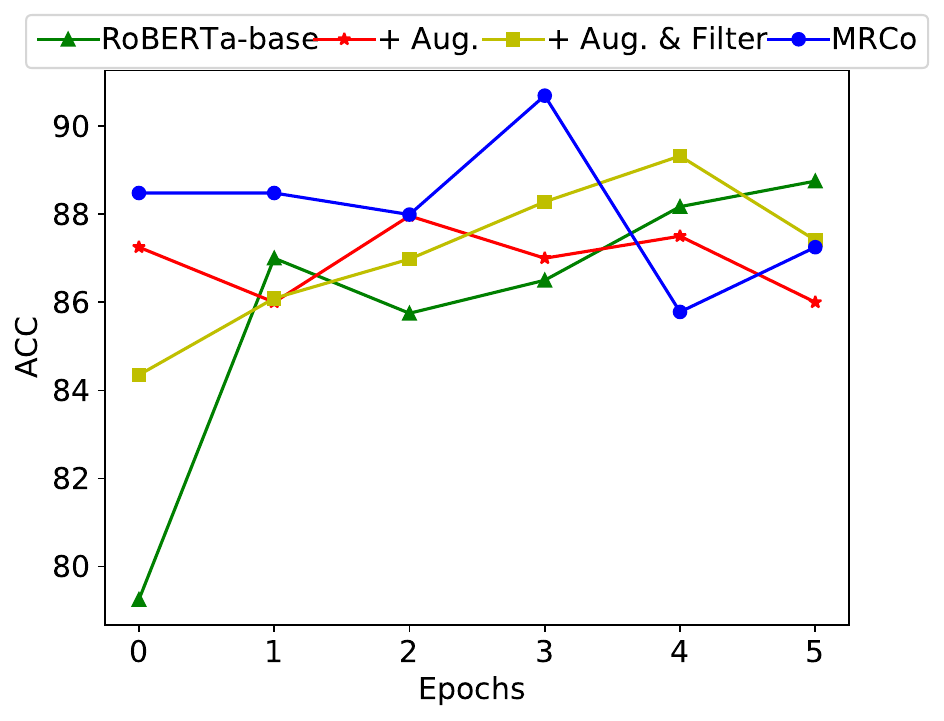}
		\caption{MRPC}
		
	\end{subfigure}
	\caption{Development set accuracy with basic encoder RoBERTa-base. }
	\label{fig:convergency_analysis}
	\vspace{-5pt}
\end{figure*}

\subsubsection{Ablation Study of the Contrastive Learning Module}
In the hyper-parameter analysis of $\lambda$, we learned that the median value of $\lambda$ achieved the best performance. To understand the impact of contrastive learning in {\model}, we assigned $\lambda=0$, indicating {\model} without contrastive learning. In Table~\ref{tab:no_contrastive}, we can observe that contrastive learning played an important role in the performance of \model. In particular, there was a 3.8\% accuracy drop without using contrastive learning compared with our original {\model}. In addition, by comparing the performance between ``\textit{+ Aug. \& Filter}" and ``{\model} \textit{w/o} Contrastive Learning", we observe that in the RTE task, only utilizing meta reweighting module is not sufficient to beat the manually designed filtering method. In addition, to understand the effectiveness of LASW dequeue-enqueue algorithm, we also considered replacing it with a FIFO algorithm (named as {\model} \textit{w/o} LASW). Such replacement downgraded our framework's performance by 1.3\% on both RTE and MRPC tasks. These observations revealed clear evidence of the non-trivial contributions of our contrastive learning design.


\subsection{Training Dynamics}
To understand how quickly our model got converged (i.e., reaching the best result), we visualized the development set accuracy with RoBERTa-base on MRPC and RTE in Fig.~\ref{fig:convergency_analysis}. Compared with the \textit{+ Aug}, which used the same amount of augmented data, our approach achieved slightly faster convergence (three epochs in {\model} vs. four epochs in \textit{+ Aug}). Different from the existing contrastive learning framework~\cite{fang2020cert}, {\model} did not require an additional training stage to learn a better feature representation at the beginning. {\model} also converged faster than the other two baselines. 

\section{Related Work}
\label{sec: relatedWork}
In this section, we summarize related work in 3 topics: (1) text augmentation; (2) meta learning and instance reweighting; and (3) contrastive learning.
\subsection{Text Augmentation}
Researchers proposed various text augmentation methods. There are dictionary-based word substitutions~\cite{miller1995wordnet,ribeiro2020beyond,zang2020word}, where words are replaced with their synonyms. Other works search for similar words in certain context-free word embedding space~\cite{alzantot2018generating,wang2015s,jiao2020tinybert}. Alternatively, works leveraging contextualized embeddings in masked language models can be applied for text augmentations~\cite{garg2020bae,wu2019conditional}. Generative text augmentations also gained research interests~\cite{wullach2020towards,cao2020hategan,anaby2020not,kumar2020data}. A large number of noise injection operations (swapping/insertion/deletion/cutoff) on words/subwords/chars can also be applied for augmentations to improve a model's robustness or for creating adversarial examples~\cite{xie2017data,xie2020unsupervised,luque2019atalaya,gao2018black,li2019textbugger,pruthi2019combating,shen2020simple}. Recent works also revealed \textit{back-translation}~\cite{qu2020coda,sennrich2016improving,edunov2018understanding} and \textit{mixup}~\cite{guo2019augmenting,chen2020mixtext} which were helpful in certain NLP tasks. Combinations of these methods were also presented in several works~\cite{weizou2019eda,coulombe2018text,rizos2019augment}. Lastly, researchers developed practical tools of the aforementioned augmenting methods~\cite{ma2019nlpaug,morris2020textattack}.
%
In this paper, we utilized five representative augmentation methods as mentioned before, namely: \textit{Wordnet}~\cite{miller1995wordnet}, \textit{Easydata}~\cite{weizou2019eda}, \textit{Checklist}~\cite{ribeiro2020beyond}, \textit{Embedding}~\cite{alzantot2018generating} and \textit{Charswap}~\cite{li2019textbugger}.

\subsection{Meta Learning and Instance reweighting}
Meta Learning~\cite{vilalta2002perspective} is also known as ``learning to learn''~\cite{lake2015human}. In this work, we mainly focus on leveraging meta learning techniques for reweighting augmented samples to reduce the noise within these augmented samples and preventing the model from collapse through bilevel optimization.


Recent work associated with data augmentations and meta learning~\cite{tang2020online,rajendran2020meta,hu2019learning,zheng2021meta,pham2021meta,ni2020data,shu2019meta} mainly focused on computer vision domain, in which the instance space of images is continuous. However, text is different from images in terms of the discrete nature of text strings. Alternatively, researchers leveraged reinforcement learning architectures where rewards' gradients could be calculated and back-propagated through the networks. Some networks also required the gradients to be back-propagated to the augmentation network. This approach is not applied to many heuristic text-based methods.
Unlike the prior works, we proposed a framework that treats augmentation techniques as black boxes, and does not require the design of any reinforcement learning rewards.

Yi~\cite{yi2021reweighting} proposed a reweight module to assign large weights to augmented samples with large loss values so that the model could pay more attention to harder-to-learn examples. Shu~\cite{shu2019meta} also utilized meta learning to learn a reweight module based on the loss of augmented samples. 
However, Shu's reweight module assigns large weights to the samples with small loss as they view large loss as an indication of the potential noise in the samples.
In this paper, we consider the quality of samples at different levels: the feature representations, and the associated label, which contain richer information than the singular-valued loss.

\subsection{Contrastive Learning}
The basic idea of contrastive learning is to encourage the feature representations of ``similar'' instances to be close while enforcing ``different'' instances to be apart. 
In unsupervised scenarios, \textit{similar instances} can be images and their augmentation variations~\cite{he2020momentum,qu2020coda}. While in supervised scenarios, such requirements are sometimes relaxed, where \textit{similar instances} can be defined as ones with the same labels~\cite{khosla2020supervised,gunel2021supervised}.

We applied contrastive learning differently, as we
defined similar augmented samples as those with similar weights generated by the meta reweighting module while large weighted augmented samples and small weighted augmented samples were naturally dissimilar. Those small weighted augmented samples were further ordered in a contrastive learning queue, with a novel proposed dequeue-enqueue algorithm rather than ordinary FIFO queue. Thus, our contrastive learning component was well adapted to the meta reweighting module. 

\section{Conclusion}
\label{sec: conclusion}
In this paper, we proposed a \textit{Meta Reweighting Contrastive learning framework~(\model)}  to reduce and exploit the noise from data augmentation.   
Our model achieved performance improvements across various augmentation strategies and backbone encoder modules in the experiments. We further conducted extensive analysis of the model components to show their non-trivial contributions and give more insights in terms of the impact of varying values of different hyper-parameters and the actual training dynamics of our framework. In the future, we plan to extend our design to broader domains such as computer vision. We are sharing our implementations for better reproducibility.

\section*{Acknowledgment}
\noindent This work was supported in part by NSF grant CNS-1755536. 
Any opinions, findings and conclusions or recommendations expressed in this material are the author(s) and do not necessarily reflect those of the sponsors.

\bibliographystyle{./bibliography/IEEEtran}
\bibliography{./bibliography/IEEEabrv,./bibliography/IEEEexample}

\end{document}